\documentclass[runningheads]{llncs}
\usepackage[T1]{fontenc}
\usepackage{graphicx}
\usepackage{booktabs}
\usepackage[misc]{ifsym}
\usepackage{xcolor}
\usepackage{float}
\usepackage{url}
\usepackage{booktabs}
\usepackage{tabularx}
\newcolumntype{C}{>{\centering\arraybackslash}X}
\newcolumntype{R}{>{\raggedleft\arraybackslash}X}
\newcolumntype{L}{>{\raggedright\arraybackslash}X}
\newcommand{\corr}{(\Letter)}
\usepackage{mwe}
\begin{document}

\title{Employing Two-Dimensional Word Embedding for Difficult Tabular Data Stream Classification}

\titlerunning{Employing 2D Word Embedding for Difficult Data Stream Classification}
\author{Author information scrubbed for double-blind reviewing}
\author{Paweł Zyblewski \corr \orcidID{0000-0002-4224-6709}}

\authorrunning{Paweł Zyblewski}

\institute{Department of Systems and Computer Networks, 
            Faculty of Information and Communication Technology, 
            Wrocław University of Science and Technology,
            Wybrzeże Wyspiańskiego 27, 50-370 Wrocław, Poland\\ \email{pawel.zyblewski@pwr.edu.pl}}

\maketitle

\begin{abstract}
Rapid technological advances are inherently linked to the increased amount of data, a substantial portion of which can be interpreted as data stream, capable of exhibiting the phenomenon of concept drift and having a high imbalance ratio. Consequently, developing new approaches to classifying difficult data streams is a rapidly growing research area. At the same time, the proliferation of deep learning and transfer learning, as well as the success of convolutional neural networks in computer vision tasks, have contributed to the emergence of a new research trend, namely Multi-Dimensional Encoding (MDE), focusing on transforming tabular data into a homogeneous form of a discrete digital signal. This paper proposes \emph{Streaming Super Tabular Machine Learning} (SSTML), thereby exploring for the first time the potential of MDE in the difficult data stream classification task. SSTML encodes consecutive data chunks into an image representation using the STML algorithm and then performs a single ResNet-18 training epoch. Experiments conducted on synthetic and real data streams have demonstrated the ability of SSTML to achieve classification quality statistically significantly superior to state-of-the-art algorithms while maintaining comparable processing time.

\keywords{Data stream  \and Multi-Dimensional Encoding \and Concept drift \and Imbalanced data.}
\end{abstract}

\section{Introduction}
The widespread access to the internet and the development of technology related to all areas of daily life resulted in the largest-ever influx of data, the analysis of which can bring both research and commercial benefits. At the same time, it is important to note that the data is now often no longer static but rather a potentially infinite stream whose characteristics can change over time \cite{bahri2021data}. Despite the increasing amount of multimodal data, often focusing on discrete digital signals, acoustic signals, or text, heterogeneous tabular data is still the most prevalent data type \cite{shwartz2022tabular}. It is common, e.g., in healthcare \cite{batko2022use}, user recommendation systems \cite{zhang2021neural}, or social media.

Analyzing streaming data is associated with computational and time complexity limitations due to the inability to store the entire stream and the need to process each instance or batch just once. In addition, depending on the application, a data stream classification system may be required to make relatively fast decisions \cite{gama2013evaluating}. The most distinctive property of the data stream is the possibility of concept drift, which, in the case of real drift, leads to a shift in the problem's decision boundary and, thus, to the degeneration of pattern recognition models. Real data streams are also often characterized by a high imbalance ratio, which additionally introduces a bias in the favor of the the majority class \cite{aguiar2023survey}. We may also have to deal with dynamic changes in the prior probability, in which the imbalance ratio changes over time, even considering the possibility of the minority class transitioning into the majority class. Therefore, when analyzing this type of data, it is necessary to use algorithms capable of constantly updating knowledge to adapt to changes occurring in the stream. One of the most popular research directions here is related to the algorithms based on classifier ensembles, combined with data preprocessing techniques \cite{krawczyk2017ensemble}.

At the same time, we have been observing the dynamic development of deep learning, whose successes in the area of computer vision have inspired the research community to replicate this success for the tabular data \cite{kadra2021well}. This is due to the heterogeneous nature of tabular data, which may consist of numerical and categorical features that are less correlated than in the case of homogeneous image, audio, or text \cite{borisov2022deep} data. The taxonomy of deep learning for tabular data distinguishes three approaches: \textbf{(i)} Data Transformation Methods, divided into Single-Dimensional and Multi-Dimensional encoding, transforming numerical and categorical features to facilitate information extraction by the neural network; \textbf{(ii)} Specialized Architectures, in which we can distinguish Hybrid Model and Transformer-based Models, proposing various network architectures to deal with tabular data, and \textbf{(iii)} Regularization Models, employing regularization schemes to alleviate data nonlinearity and complexity.

Despite the existence of deep neural network-based solutions dedicated to the classification of tabular data, they are still not the preferred solution for the classification of data streams, and more emphasis is placed on developing less complicated algorithms based on classic, non-deep pattern recognition models such as decision trees \cite {manapragada2018extremely}. This is due to limited access to data and constraints in computational complexity, which translates into difficulties in model selection and optimizing hyperparameters. Adapting neural networks to process data streams is gradually being undertaken in the literature \cite{duda2020training}. However, it is still considered one of the research challenges in the area of deep learning \cite{borisov2022deep}.

The purpose of this paper is to combine two research areas that are rarely grouped together, namely the classification of imbalanced data streams with concept drift and the application of deep learning in tabular data classification. To this end, the \emph{Streaming Super Tabular Machine Learning} (SSTML) approach, based on the \emph{Super Tabular Machine Learning} (STML) method developed by Sun et al. \cite{sun2019supertml}, is proposed. STML employs two-dimensional word embedding to store the values found in the feature vector of an instance in a black-and-white image. This representation is then passed to the ResNet18 architecture to classify difficult data streams with various characteristics. The experimental evaluation aims to make plausible the hypothesis that \emph{using multi-dimensional encoding combined with deep convolutional networks can achieve classification quality surpassing state-of-the-art ensemble algorithms for imbalanced tabular data stream classification while maintaining time complexity acceptable for batch processing}.

In brief, the main contributions of this article are as follows:
\begin{itemize}
    \item Proposition of \emph{Streaming Super Tabular Machine Learning} (SSTML) approach, applying the MDE approach to difficult tabular data stream classification for the first time.
    \item Evaluation of SSTML on synthetic, semi-synthetic, and real data streams in terms of classification quality, handling concept drift, and processing time.
    \item Comparison of SSTML to state-of-the-art ensemble data stream classification algorithms.
\end{itemize}

\section{Related Works}
The following section briefly describes the three research areas most relevant to this work, i.e., \textbf{(i)} imbalanced data stream classification, \textbf{(ii)} multi-dimensional encoding, and \textbf{(iii)} neural networks for data stream analysis.

\subsection{Classifier ensemble for difficult data stream}
Despite almost three decades of research and many algorithms currently available, classifying imbalanced data streams with concept drift is still one of the leading areas of machine learning. Stream data analysis algorithms can be designed for online processing, where processing occurs instance by instance or batch processing, in which the stream is processed in disjoint windows. This work focuses on batch processing, which, due to the more extensive training set, may enable better generalization ability in the current concept but is associated with a delayed response obtained due to waiting for the next chunk to be available \cite{aguiar2023survey}.

Data imbalance is a problem commonly encountered in data streams where the imbalance ratio may be static or dynamic \cite{aminian2019study}. Most real streams do not have a fixed imbalance ratio, and their characteristics may change frequently \cite{wang2018systematic}. Therefore, data stream classification algorithms should achieve high classification quality regardless of class distribution, while most algorithms designed for imbalanced data streams do not achieve satisfactory results when the class sizes are comparable. At the same time, algorithms dedicated to balanced data often have problems with the correct classification of streams with skewed class distribution, and only some of the state-of-the-art proposals can deal with both cases \cite{cano2020kappa}. Methods for dealing with imbalances are divided into two groups: \textbf{(i)} data-level approaches and \textbf{(ii)} algorithm-level approaches \cite{aguiar2023survey}. Data-level approaches focus on preprocessing data -- often through oversampling or undersampling-- to change its characteristics before the classification process, while Algorithm-level approaches focus on modifying classification algorithms' training process to reduce majority class bias resulting from skewed distribution.

The most common approaches to imbalanced data stream classification are based on classifier ensemble combined with data preprocessing techniques. By ensuring diversity, updating the pool of available models, and their appropriate combination, it is possible to improve the predictive ability and achieve fluid adaptation to changes that may occur in the stream over time \cite{brzezinski2018ensemble}. Among the proven proposals, we can distinguish here \emph{Learn++.\textsc{cds}} (Concept Drift with Smote) and \emph{Learn++.\textsc{nie}} (Nonstationary and Imbalanced Environments) by Ditzler and Polikar \cite{Ditzler:2013}. \emph{Learn++.\textsc{cds}} extends the \emph{Learn++.\textsc{nse}} (Non-Stationary Environments) algorithm by using the SMOTE technique to balance class sizes, while \emph{Learn++.\textsc{nie}} employs a different penalty constraint to balance predictive accuracy on all classes and employs a bagging-based sub-ensemble instead of oversampling. In the case of online data stream classification, the Oversampling Online Bgging (OOB) and Undersampling Online Bagging (UOB) algorithms developed by Wang et al. and extending Online Bagging by modifying the value of the $\lambda$ parameter for the Poisson distribution depending on the current degree of class imbalance in the data stream \cite{Wang2015}. Klikowski and Woźniak employed one-class classifiers trained on clustered data \cite{klikowski2020employing}. Zyblewski et al. coupled Dynamic Classifier Selection with preprocessing in order to classify imbalanced data streams \cite{zyblewski2021preprocessed}. Cano and Krawczyk proposed \emph{Kappa Updated Ensemble} (\textsc{kue}) \cite{cano2020kappa} combining batch-based and online approaches. \textsc{kue} dynamically weights and selects base classifiers using the Kappa statistic. The same authors introduced \emph{Robust Online Self-Adjusting Ensemble} (\textsc{rose}) for non-stationary data stream classification \cite{cano2022rose}, which trains online learners based on data views to ensure high diversity of classifier pool. It also uses drift detectors to respond rapidly to changes in the data distribution while proposing effective techniques for handling data imbalance. Wozniak et al. proposed \cite{wozniak2023active} employing various built-in mechanisms (including, e.g., classifier weighting and aging) in order to construct a self-updating classifier pool capable of changing its lineup to react to concept drift and data imbalance.

\subsection{Multi-Dimensional Encoding}
As confirmed by numerous scientific publications, convolutional networks are successfully used for image, video and audio classification \cite{satt2017efficient}. Deep networks also enable transfer learning, allowing models to apply previously acquired knowledge to the problem at hand \cite{zhuang2020comprehensive}. 

\emph{Multi-Dimensional Encoding} \cite{borisov2022deep} attempts to transform tabular data into a more homogeneous form of digital discrete signals. One example here is the \emph{Super Tabular Machine Learning} (STML) proposed by Sun et al. \cite{sun2019supertml}, which transforms tabular data into an image by rewriting the feature values of a given problem instance onto it. Another approach is presented by the \emph{Image Generator for Tabular Data} (IGTD) \cite{zhu2021converting} developed by Zhu et al.  The algorithm searches for an optimized mapping by minimizing the difference between the ranking of distances between features and the ranking of distances between their associated pixels in the image. Damri et al. proposed the \emph{Feature Clustering-Visualization} (FC-Viz) \cite{damri2023towards}, which transforms each instance of tabular data into a 2D pixel-based representation, where pixels representing highly correlated and interacting features are adjacent to each other. 

\begin{figure}[!htb]
    \centering
    \includegraphics[width=.75\textwidth,trim={0.0cm 2.0cm 0.0cm 1.0cm},clip]{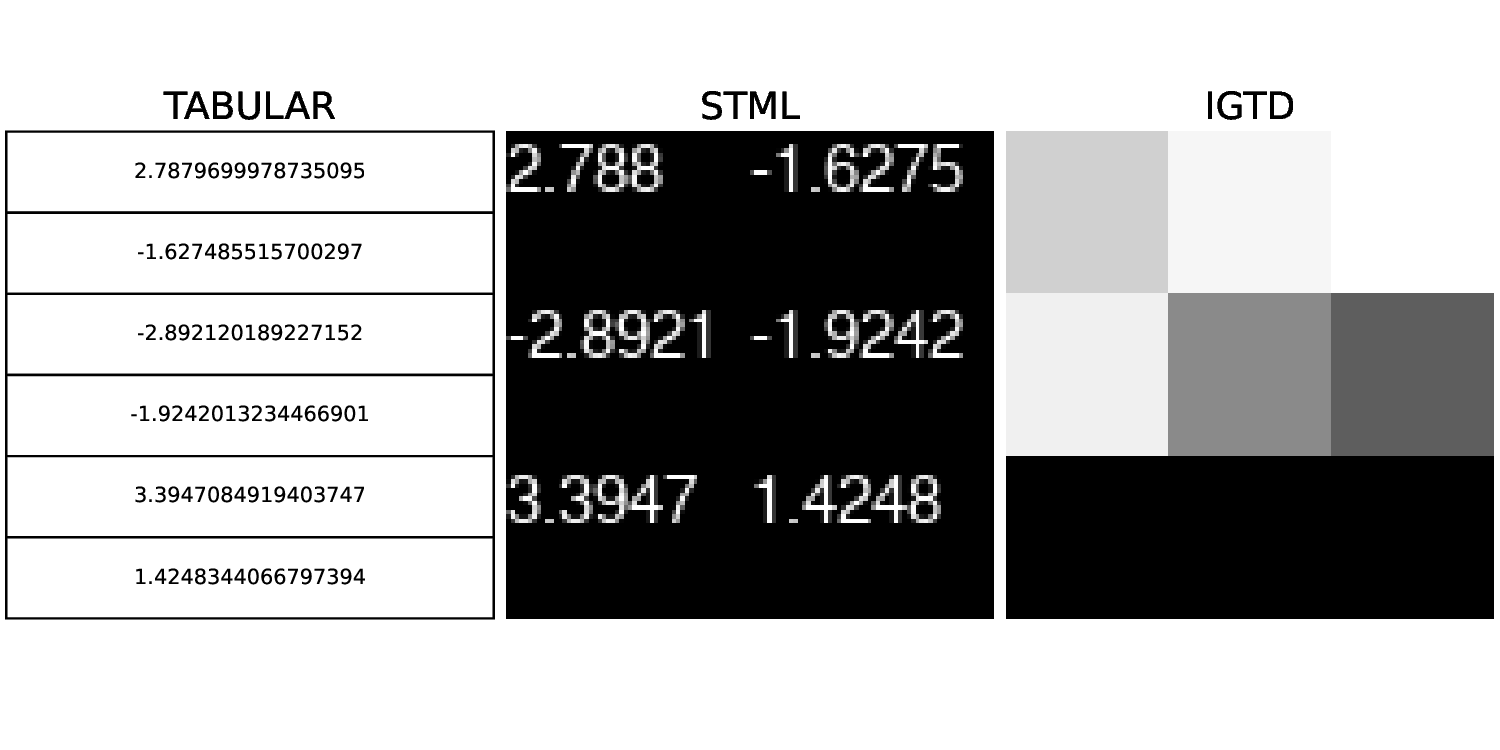}
    \caption{An example of encoding a single instance of a synthetically generated tabular problem into a two-dimensional discrete digital signal using STML and IGTD techniques.}
    \label{fig:encoding}
\end{figure}

\subsection{Neural networks for tabular data stream analysis}
As mentioned earlier, despite their success in many pattern recognition tasks, neural networks are rarely used for data streams, while methods with lower complexity are preferred \cite{borisov2022deep}. However, new works are constantly emerging, examining various areas of application of neural networks for streaming data. Sahoo et al. proposed a novel Hedge Backpropagation (HBP) method for effectively updating deep neural network parameters \cite{sahoo2017online}. Duda et al. introduced and analyzed two methods for neural network training on streaming data: Boosting-based and Bagging-based training with drift detector \cite{duda2020training}. Haug et al. proposed a FIRES framework for online feature selection by treating neural network model parameters as random variables and penalizing features with high uncertainty \cite{haug2020leveraging}. The experiments that were conducted showed that the model complexity did not affect the quality of the obtained feature sets. Additionally, Haug and Kasneci used a similar approach to detect concept drifts based on model parameter values using the ERICS framework \cite{haug2021learning}. Guzy and Woźniak demonstrated the usefulness of appropriate dropout regularization use in the event of recurring concept drift occurrence \cite{guzy2020employing}.

At the same time, with direct reference to the topic of this article, no works were found that attempted to apply any form of Multi-Dimensional Encoding for the imbalanced drifting data stream classification task. He and Sayadi employed such methods for malware detection in the Internet of Medical Things (IoMT) \cite{he2023image}. While this work was based on streaming data, it did not address concept drift or imbalance. There were also no comparisons of the proposed approaches with state-of-the-art imbalanced data stream ensemble classification algorithms. In addition, Basu et al. predicted rare fluctuations in financial time series using Siamese-type neural networks. The suggested approach stores input features as tabular data in a sliding window generates an image of the time-series window, and predicts rare occurrences using a pre-trained convolutional neural network \cite{basu2022novel}.

\section{Streaming Super Tabular Machine Learning}
As the above literature review indicates, deep learning-based solutions are often discarded in favor of less complex algorithms despite the potential of multi-dimensional encoding and neural networks and the ongoing need for new approaches to classifying difficult data streams. For this reason, this paper proposes \emph{Streaming Super Tabular Machine Learning} (SSTML) to fill this niche and make a first step in exploring the potential of multi-dimensional encoding in dealing with difficult data streams.

The main assumption when developing SSTML was to keep the computational complexity as low as possible while using the potential of deep learning for tabular data. Therefore, we consider only the batch-based stream processing scenario, where the need to wait for subsequent data chunks alleviates potential problems that may result from the increased processing time.

\begin{figure}[!htb]
  \begin{center}
    \includegraphics[width=0.99\textwidth]{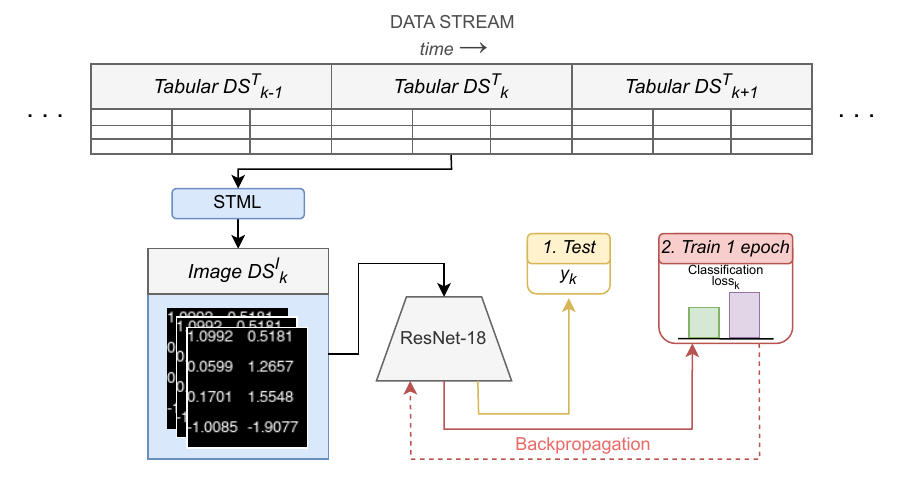}
  \end{center}
  \caption{The general scheme of the proposed SSTML approach.}
  \label{fig:sstml}
\end{figure}

From the pool of multi-dimensional encoding techniques, STML was chosen for its encoding speed and ease of implementation while offering improved classification quality \cite{leon2021dengue}. These features make STML a natural choice for initial research related to data stream classification. Due to its widespread use and relatively small size, the ResNet-18 architecture was used as a convolutional network. In addition, to ensure low processing time, only one learning epoch is carried out in each data chunk. The batch size was set to $8$, and an SGD optimizer with a learning rate of $0.001$ and momentum of $0.9$ was applied. Cross-entropy loss was chosen as the loss function with weights depending on the degree of imbalance of the problem.

The data stream processing using SSTML is shown in Fig. \ref{fig:sstml}. We treat the data stream as a series of tabular data chunks $DS^T_k$ with a predetermined size of $N$, where $k$ denotes the batch index. Each incoming data chunk is encoded using STML into a set of two-dimensional discrete digital signals $DS^I_k$ containing $N$ square images with a side of defined size. Each of the $N$ images belonging to $DS^I_k$ is then duplicated three times to obtain an image representation containing three channels for the ResNet-18 architecture. Then, for each data chunk $DS^I_k$, ResNet-18 performs inference and one training epoch, according to the Test-Then-Train protocol.  

\begin{figure}[!htb]
  \begin{center}
    \includegraphics[width=0.99\textwidth]{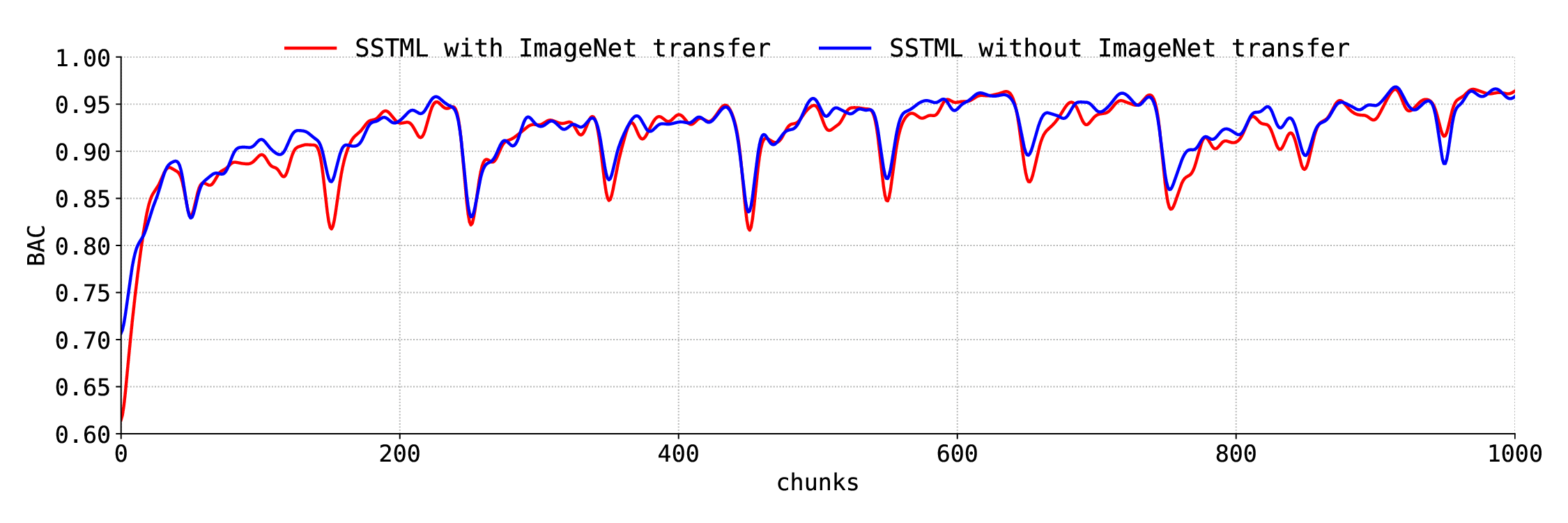}
  \end{center}
  \caption{Comparison of SSTML behavior with and without transfer learning. A Gaussian filter was applied for visualization purposes.}
  \label{fig:transfer}
\end{figure}

Many works related to computer vision use transfer learning as an implicit solution without considering the possibility of negative transfer. Therefore, prior to the study, a preliminary experiment was conducted to determine the validity of using the ResNet-18 pre-trained on the ImageNet dataset. A single stream generated using the stream-learn library \cite{ksieniewicz2022stream} was used, containing $3000$ chunks of $250$ instances, $30$ sudden drifts, $10\%$ minority class, $1\%$ label noise, and $8$ informative features. The differences are relatively small, as shown in Fig. \Ref{fig:transfer}. However, SSTML without pretreated weights has slightly higher balanced accuracy in places and reports smaller drops when concept drift occurs. SSTML without knowledge transfer also starts from a higher BAC level, indicating a low translation of the ImageNet set into the STML representation and potentially negative transfer. Therefore, ResNet-18 without pre-trained weights was used for the study.

\section{Experimental Evaluation}
This section describes the details of the experimental study conducted to investigate the properties of SSTML. The experiments conducted were designed to answer the following research questions:
\begin{enumerate}
    \item Is SSTML capable of achieving classification quality superior to state-of-the-art ensemble algorithms dedicated to difficult data streams?
    \item How does STML behave compared to ensemble algorithms in the event of concept drift occurence?
    \item Does processing time of SSTML enable its use in the real data stream classification tasks?
\end{enumerate}

\subsection{Set-up}
The research conducted was planned taking into account the following components:\\
\textbf{Data} The study was conducted on 130 streams in total, including three generators and two real streams:
\begin{itemize}
    \item stream-learn synthetic -- \textbf{chunks:} 3000, \textbf{chunk size:} 250, \textbf{\#drifts:} 30, \textbf{drift type:} sudden, gradual and incremental, \textbf{recurring:} no, \textbf{imbalance:} 15\% and 5\% of minority class, \textbf{label noise:} 1\% and 5\%, \textbf{\#features:} 8, \textbf{\#replications:} 5, \textbf{STML image size (px):} 50x50,
    \item stream-learn semi-synthetic -- \textbf{datasets:} ecoli-0-1-4-6-vs-5, glass5, popfailures, spectfheart, and yeast6, \textbf{chunks:} 2000, \textbf{chunk size:} 250, \textbf{\#drifts:} 20, \textbf{drift type:} sudden and incremental, \textbf{recurring:} no, \textbf{\#replications:} 5, \textbf{image size (px):} depending on the dataset, \textbf{imbalance:} depending on the dataset,
    \item MOA -- \textbf{generators:} SEA, RBF, hyperplane, \textbf{chunks:} 2000, \textbf{chunk size:} 250, \textbf{\#drifts:} 6, \textbf{drift type:} sudden and incremental, \textbf{recurring:} yes, \textbf{imbalance:} 10\% and 5\% of minority class, \textbf{\#features:} depending on the generator, \textbf{\#replications:} 3, 
    \item CovType stream -- \textbf{chunks:} 265, \textbf{chunk size:} 1000, \textbf{\#features:} 54, \textbf{IR:} 4,
    \item Poker stream -- \textbf{chunks:} 359 \textbf{chunk size:} 1000, \textbf{\#features:} 10, \textbf{IR:} 10.
\end{itemize}

\textbf{Reference methods} Throughout the conducted experiments, SSTML was compared with the following approaches: 
\textbf{(i)} \textit{Hoeffding Tree}  Hellinger split criterion (HF), 
\textbf{(ii)} \textit{Learning++CDS} (CDS),
\textbf{(iii)} \textit{Learning++.NIE} (NIE),
\textbf{(iv)} \textit{Kappa Updated Ensemble} (KUE), and 
\textbf{(v)} \textit{Robust Online Self-Adjusting Ensemble} (ROSE).
HF was used as the base classifier for all ensemble methods, and the maximum classifier pool size was set to $10$. The selection of methods, base classifier, and pool size was based on the algorithm's prevalence and the literature \cite{aguiar2023survey,cano2020kappa,cano2022rose}.

\textbf{Experimental protocol} All experiments were carried out using the Test-Then-Train protocol according to stream-learn implementation to guarantee a robust experimental evaluation. The results were supported by the Wilcoxon signed-rank test with $\alpha=0.05$, allowing for a global comparison of the tested algorithms. 

\textbf{Reproducibility} All experiments were performed in Python and can be replicated using the publicly available GitHub repository \footnote{\url{https://github.com/w4k2/2d_mde_stream}}. The repository also contains additional results, not included in the work due to space limitations. The stream-learn, scikit-multiflow \cite{montiel2018scikit}, and PyTorch libraries implement state-of-the-art algorithms. Synthetic and semi-synthetic streams were generated using stream-learn and MOA \cite{MOA}.

\subsection{Experiment 1 -- Synthetic data streams}
\begin{figure}[!htb]
  \begin{center}
    \includegraphics[width=0.99\textwidth]{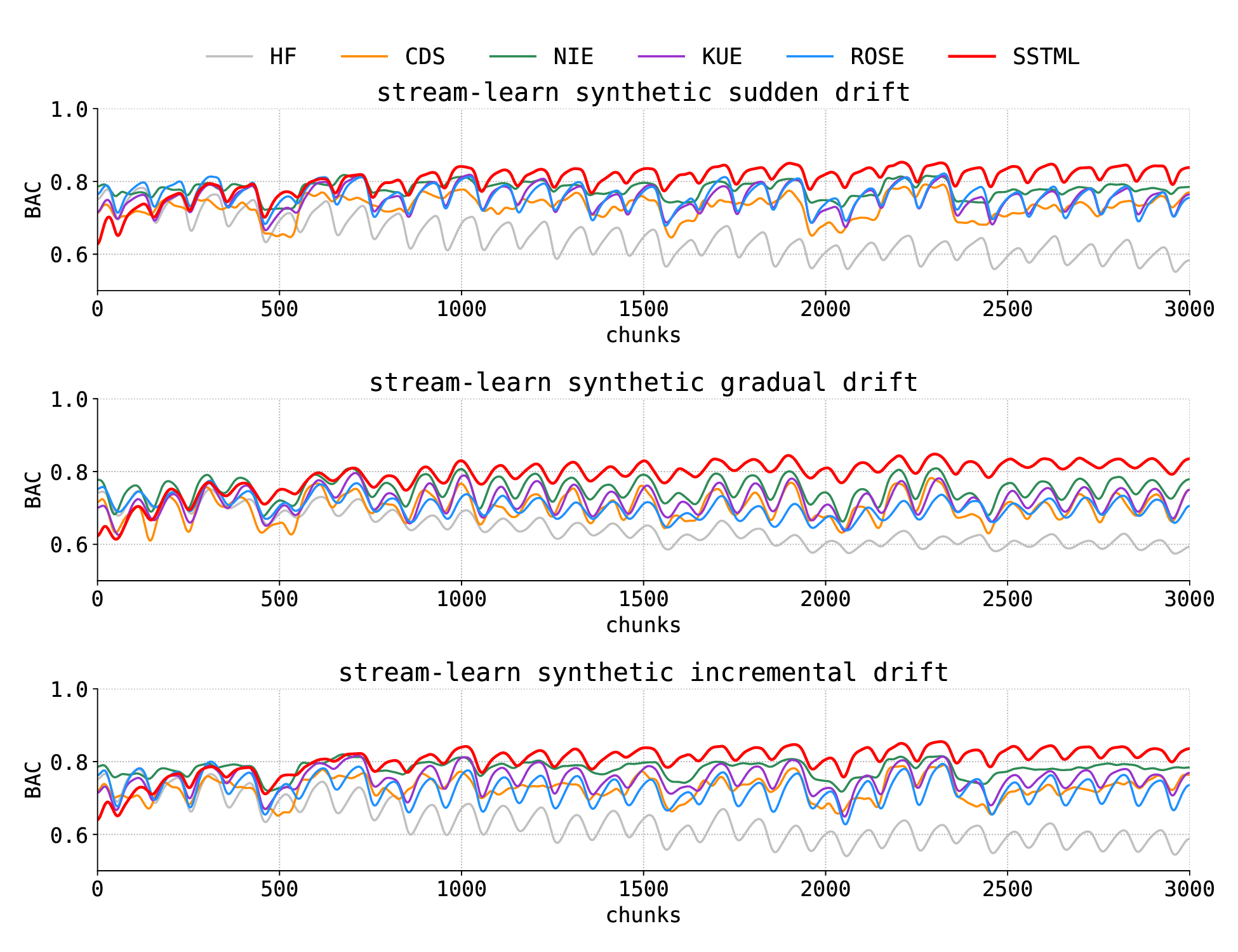}
  \end{center}
  \caption{Comparison of SSTML with reference methods in terms of BAC on synthetic streams. A Gaussian filter was applied for visualization purposes.}
  \label{fig:syn_drift1}
\end{figure}

\begin{figure}[!htb]
  \begin{center}
    \includegraphics[width=0.49\textwidth, trim={0 0 0 47em},clip]{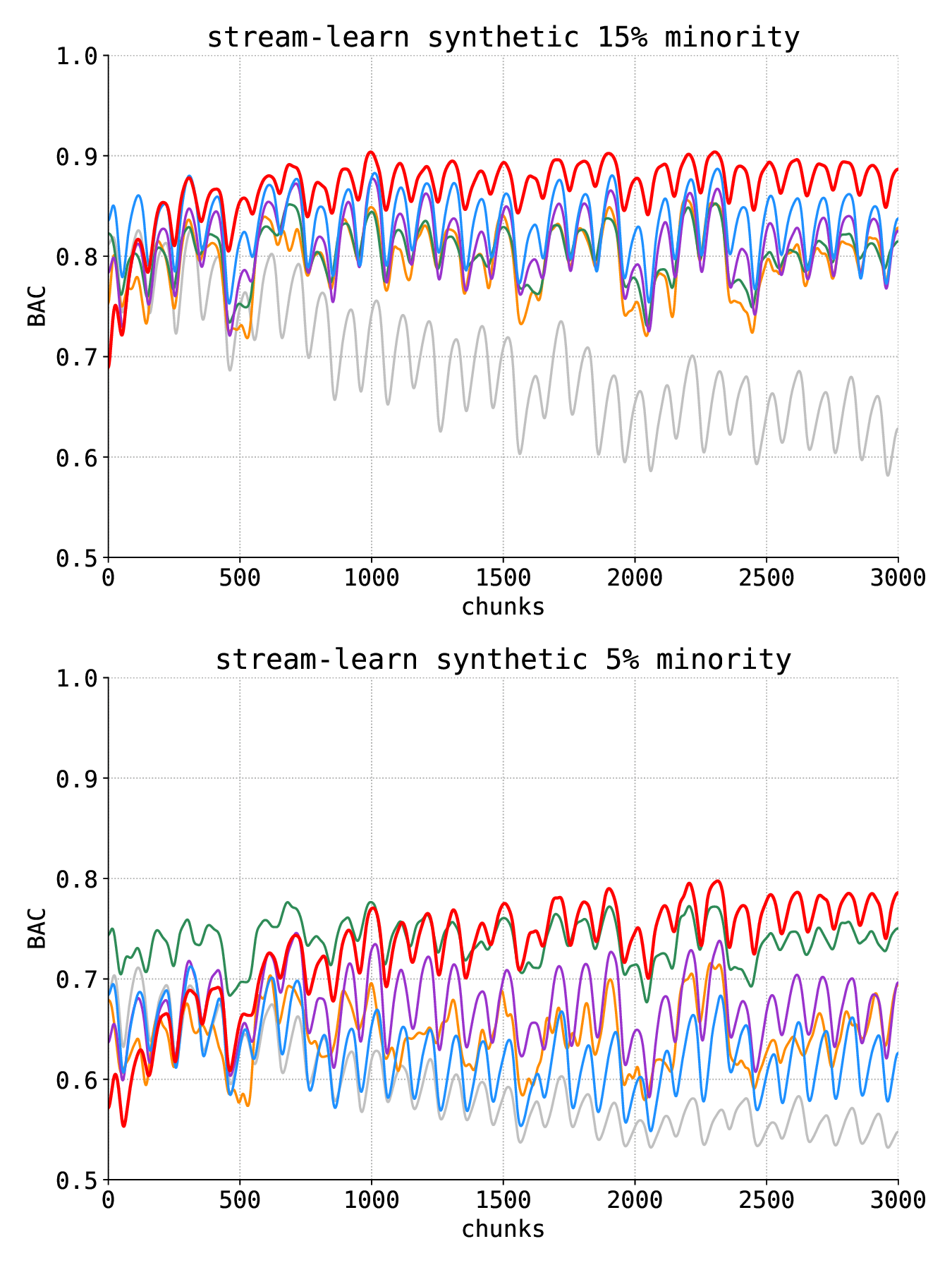}
    \includegraphics[width=0.49\textwidth, trim={0 0 0 47em},clip]{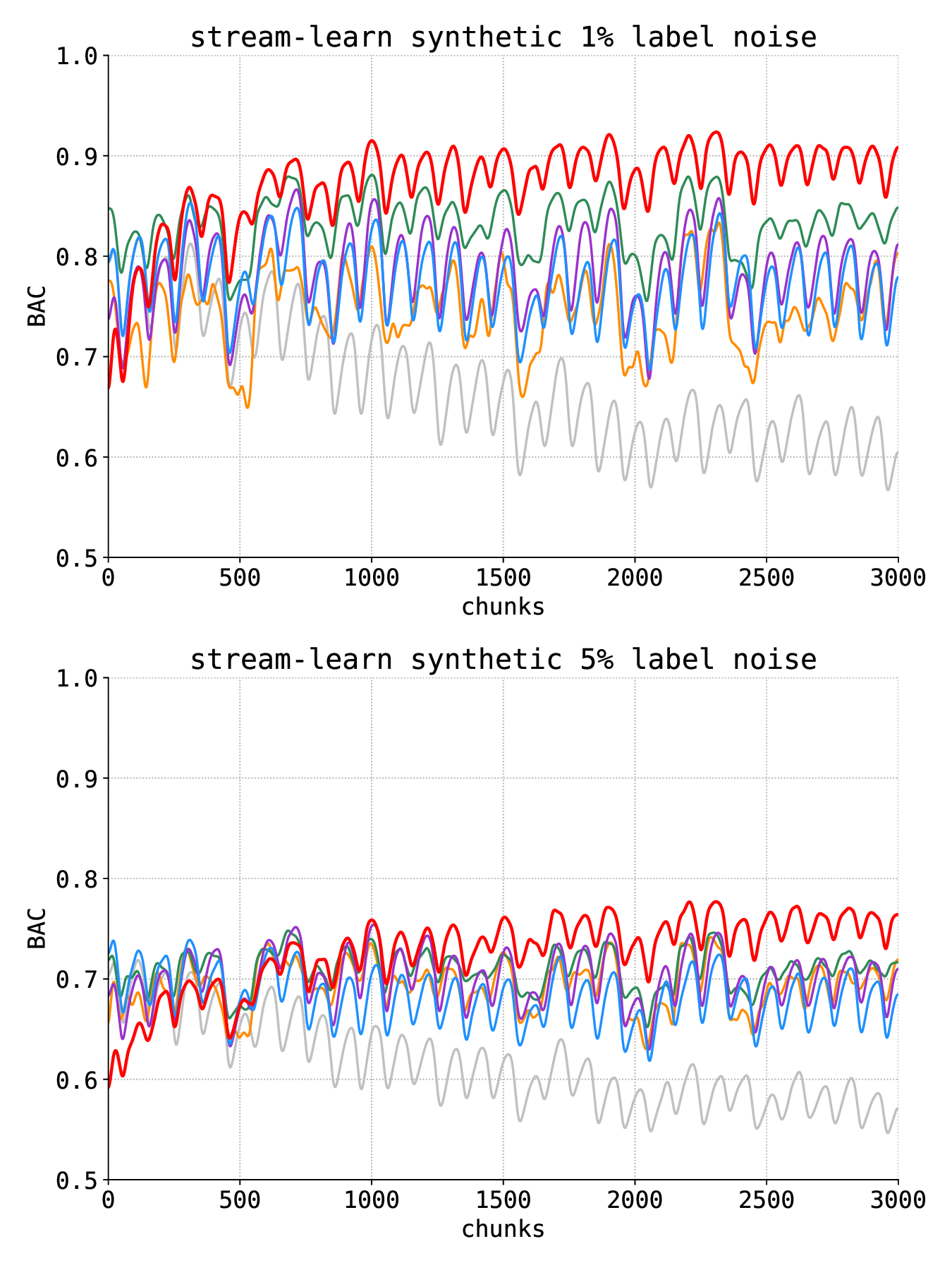}
    \includegraphics[width=0.98\textwidth]{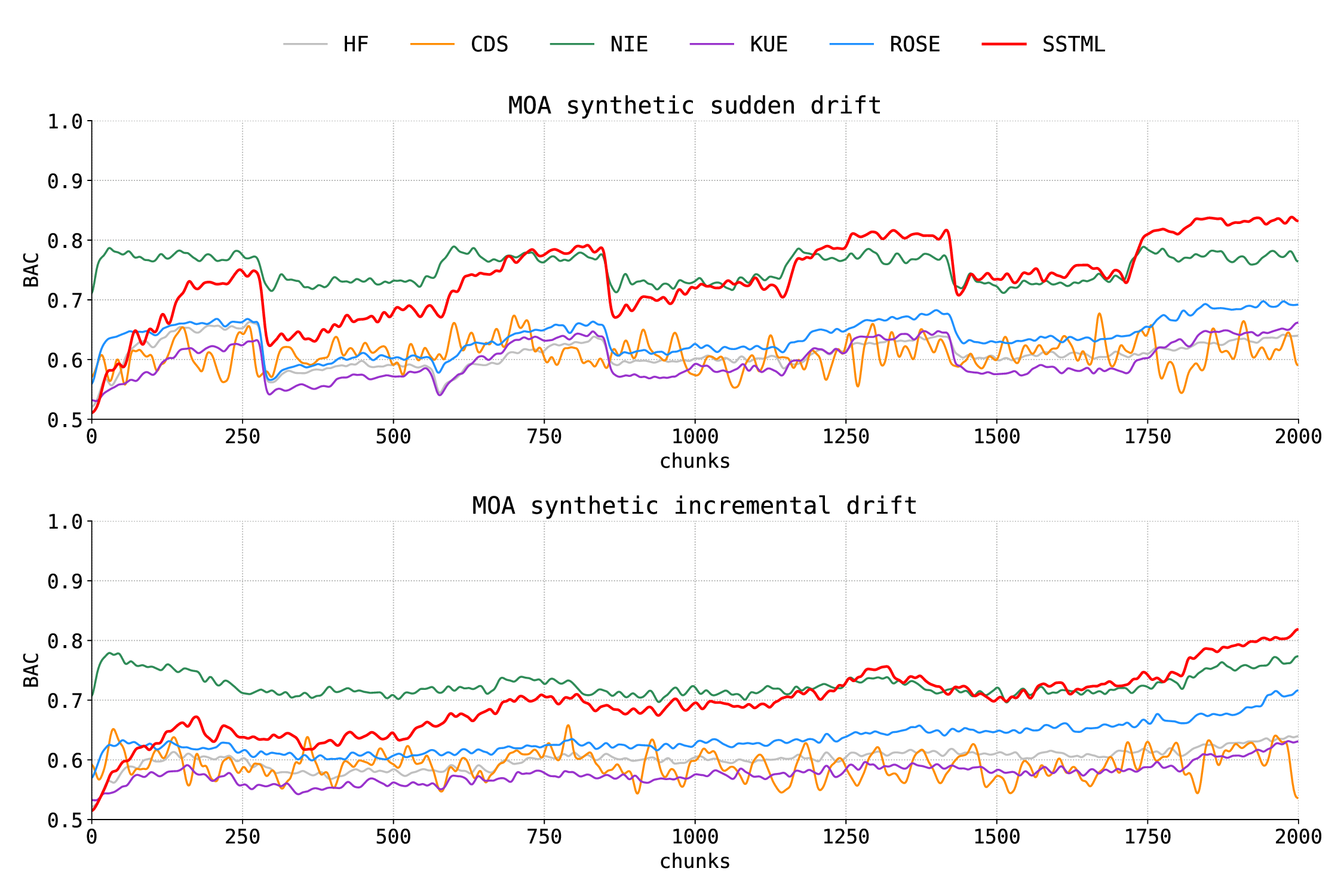}
  \end{center}
  \caption{Comparison of SSTML with reference methods in terms of BAC on synthetic streams. A Gaussian filter was applied for visualization purposes.}
  \label{fig:syn_drift2}
\end{figure}

Fig. \ref{fig:syn_drift1} and Fig. \ref{fig:syn_drift2} show the experimental evaluation results for synthetic streams in terms of balanced accuracy score (BAC). As we can observe, SSTML, regardless of the stream characteristics, starts from a level comparable to or lower than the reference methods. However, it gradually improves until it surpasses all state-of-the-art algorithms. This is due to ResNet-18's inherently higher generalization potential compared to non-deep approaches, which, even when learning is slowed down due to the use of a single epoch of training in each data chunk, still produces satisfactory results. These results demonstrate the versatility of SSTML, which retains its properties for various degrees of imbalance, label noise levels, and recurring drift in MOA streams. The best of the reference methods -- especially in the case of strong imbalance -- is NIE, which quickly achieves full generalization capability. ROSE and KUE exhibit similar properties, and their classification quality decreases significantly as the degree of imbalance increases. CDS behaves similarly to the previously mentioned methods, but it shows low stability due to the use of SMOTE and a relatively simple weighting scheme. Due to the lack of a mechanism to deal with concept drift, HF degenerates very quickly and serves only as a reference value for the other methods.

\begin{figure}[!htb]
  \begin{center}
    \includegraphics[width=0.49\textwidth]{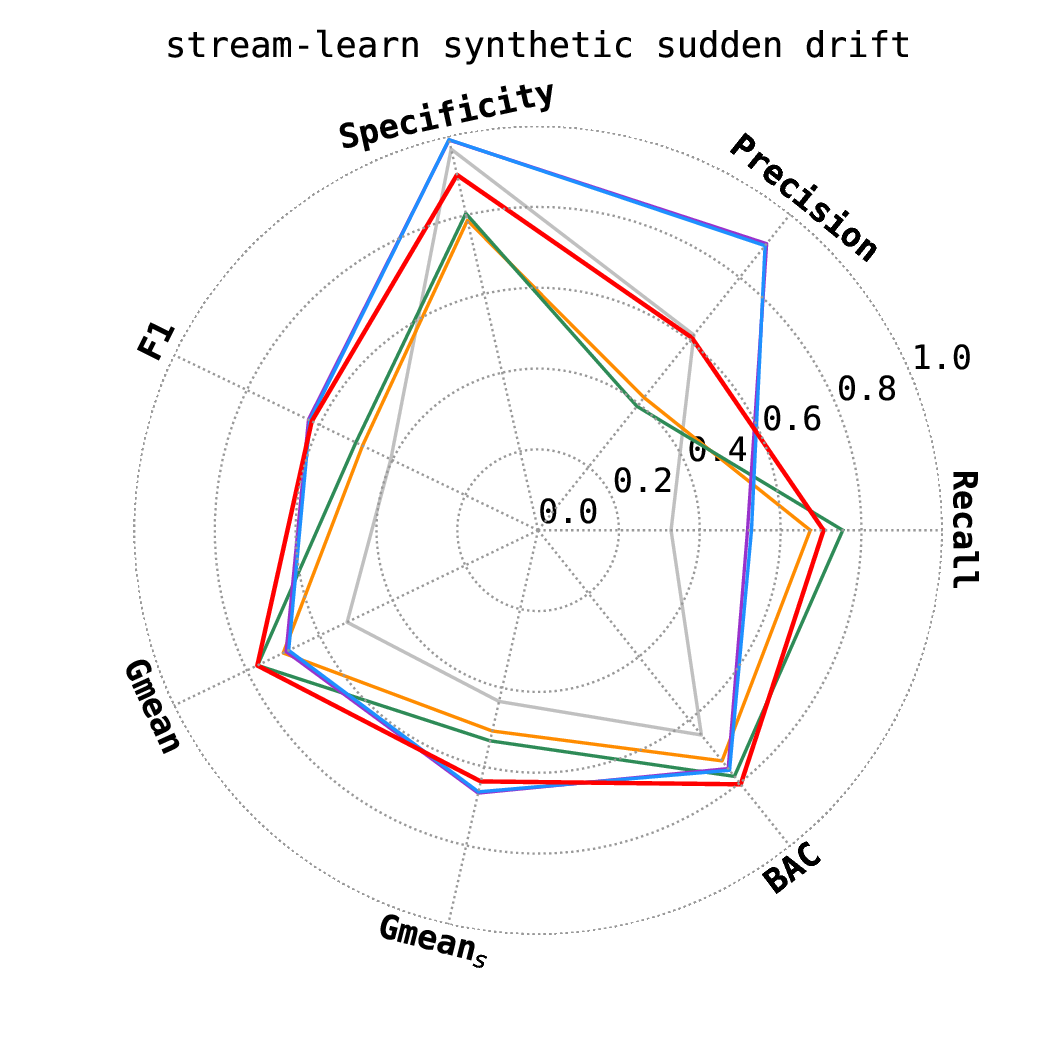}
    \includegraphics[width=0.49\textwidth]{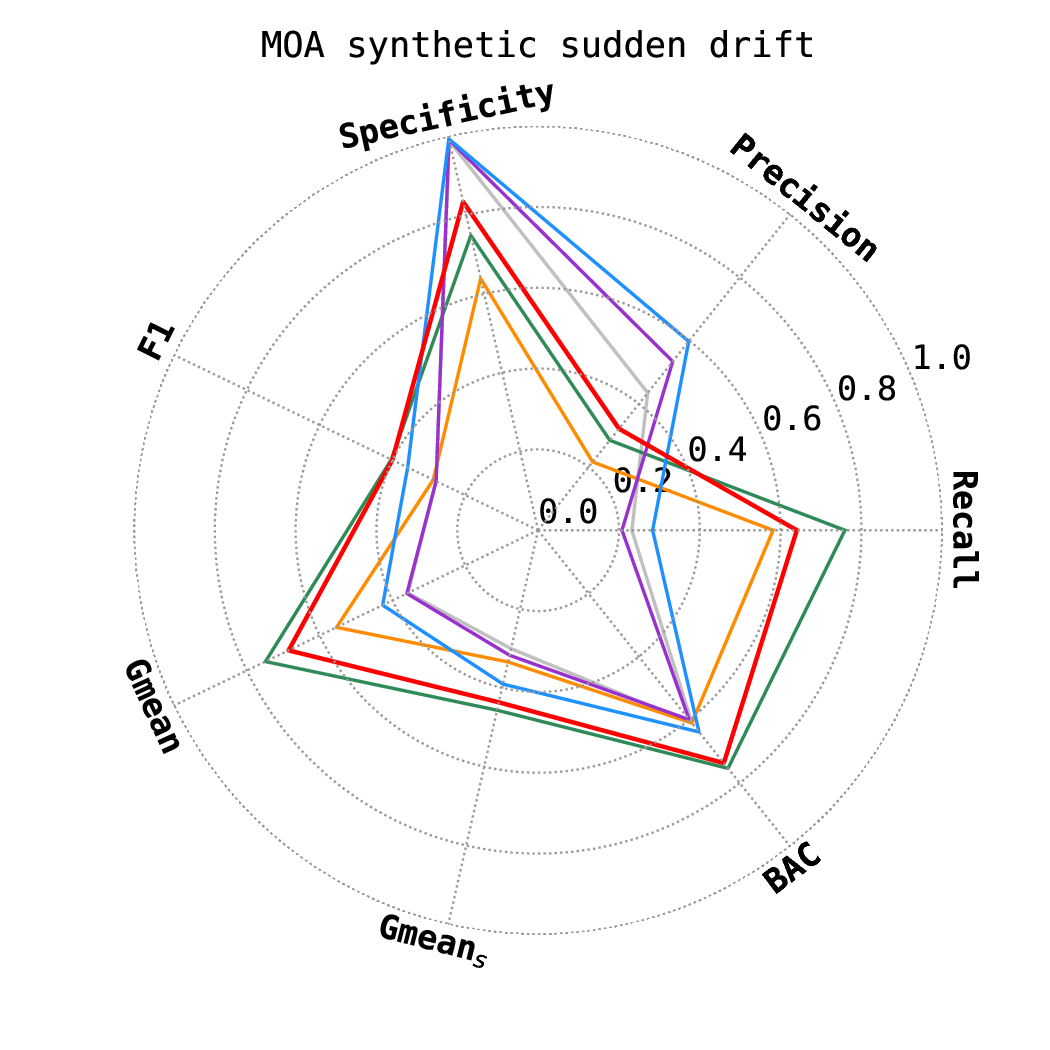}
    \includegraphics[width=0.98\textwidth, trim={0 73em 0 2em},clip]{moa_drift_bac.eps}
  \end{center}
  \caption{Comparison of SSTML with reference methods for sudden concept drift in synthetic streams in terms of averaged metrics.}
  \label{fig:radar_synth}
\end{figure}

Fig. \ref{fig:radar_synth} shows the average values of other metrics commonly used in imbalanced data classification for streams with sudden concept drifts. We see that SSTML is the most balanced method for all metrics for both generators. It has a better minority class recognition capability than ROSE and KUE while showing better precision than CDS and NIE. It is worth noting here that the entire length of the stream was averaged here, so the values presented here do not reflect SSTML's ability to improve its generalization power continuously.

Tab. \Ref{tab:wilcoxon} shows the BAC-based Wilcoxon signed-ranks statistical test results depending on the generator and concept drift type. For stream-learn, SSTML was found to be statistically significantly the best among all the methods analyzed. In the case of MOA, SSTML and NIE were the only ones to perform statistically significantly better than some of the other reference methods. Again, the analysis was conducted based on the entire stream length without taking into account the upward trend of SSTML.

\begin{table}[!htb]
    \centering
    \scriptsize
    \caption{Results of Wilcoxon statistical analysis ($\alpha=0.05$) in terms of BAC. The first row shows the average ranks, and below are the indexes of statistically significantly inferior methods.}
    \begin{tabularx}{0.99\textwidth}{lCCCCCC}
\toprule
  \textbf{Drift type}   & \textbf{HF$^{(1)}$}    &   \textbf{CDS$^{(2)}$} & \textbf{NIE$^{(3)}$}   & \textbf{KUE$^{(4)}$}   & \textbf{ROSE$^{(5)}$}   & \textbf{SSTML$^{(6)}$}   \\
\midrule
 & \multicolumn{6}{c}{\bfseries\textsc{stream-learn synthetic}}\\
 sudden       & 1.000        &          2.550 & 4.200         & 3.850         & 3.800          & 5.600           \\
              & ---          &          1    & 1, 2          & 1, 2          & 1, 2           & all             \\
 gradual      & 1.200        &          2.950 & 4.200         & 3.700         & 3.300          & 5.650           \\
              & ---          &          1    & 1, 2          & 1, 2          & 1              & all             \\
 incremental  & 1.050        &          2.7  & 4.350         & 3.850         & 3.450          & 5.600           \\
              & ---          &          1    & 1, 2          & 1, 2          & 1              & all             \\
\midrule
 & \multicolumn{6}{c}{\bfseries\textsc{moa synthetic}}\\
 sudden       & 2.444 & 3.000 & 4.444   & 2.889 & 3.667  & 4.556   \\
              & ---   & ---   & 1, 2    & ---   & ---    & 1, 2, 4 \\
 incremental  & 2.667 & 3.222 & 4.778   & 2.333 & 3.667  & 4.333   \\
              & ---   & ---   & 1, 2, 4 & ---   & ---    & 4       \\
\bottomrule
\end{tabularx}
    \label{tab:wilcoxon}
\end{table}

\subsection{Experiment 2 -- Semi-synthetic and real data streams}
\begin{figure}[!htb]
  \begin{center}
    \includegraphics[width=0.98\textwidth, trim={0 2em 0 5em},clip]{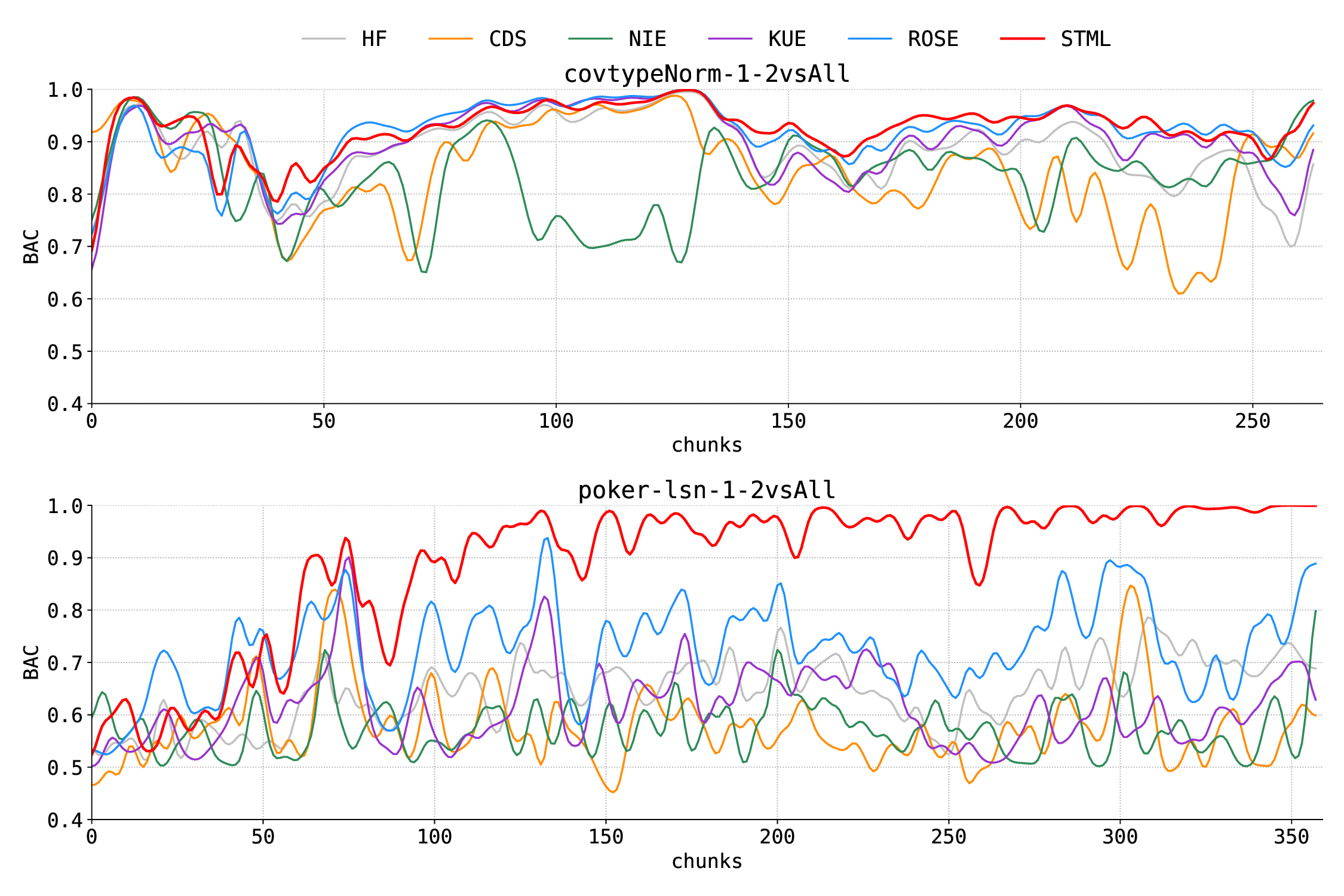}
    \includegraphics[width=0.98\textwidth, trim={0 73em 0 2em},clip]{moa_drift_bac.eps}
    \includegraphics[width=0.98\textwidth, trim={0 40em 0 0},clip]{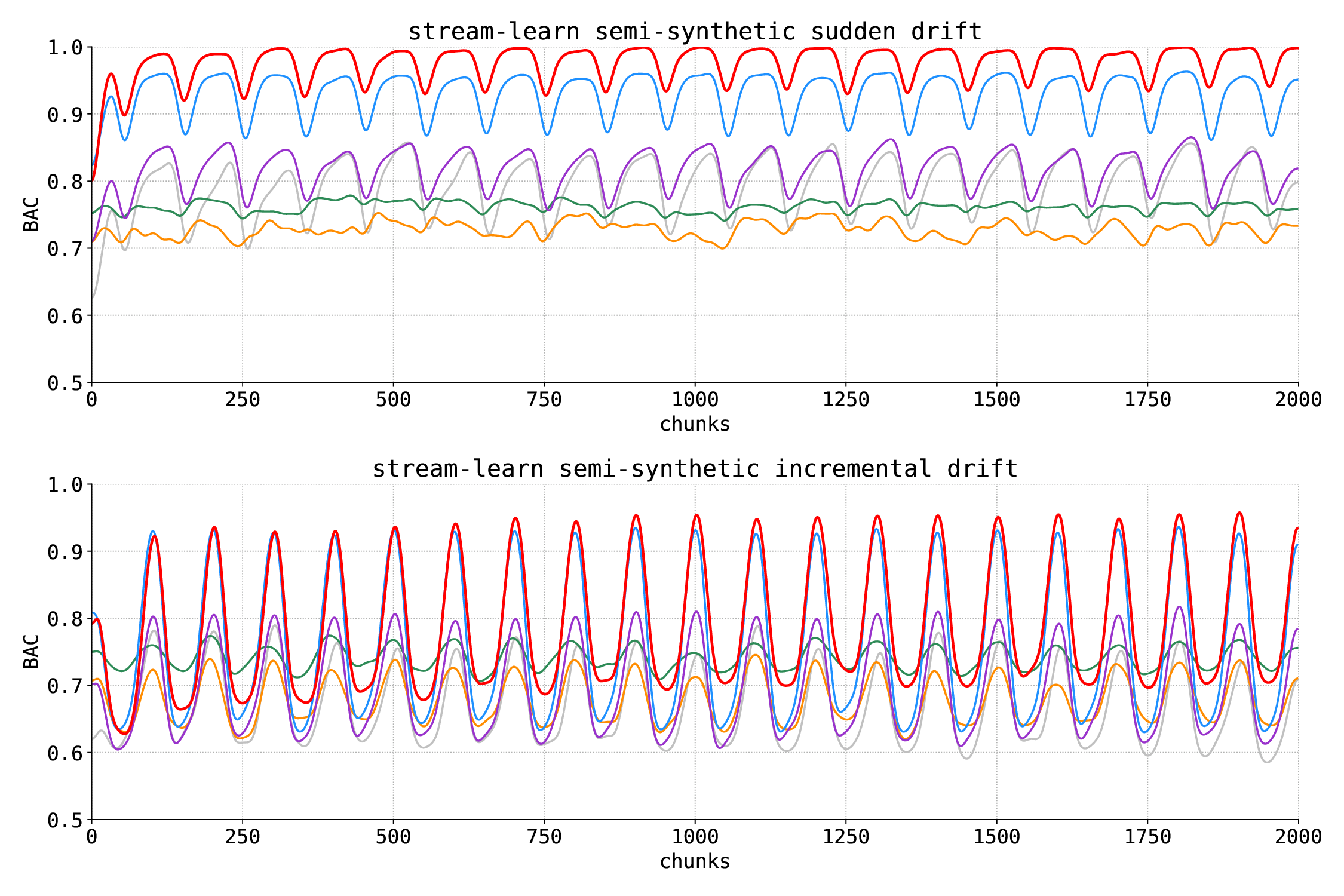}
  \end{center}
  \caption{Comparison of SSTML with reference methods in terms of BAC on real and semi-sythetic streams. A Gaussian filter was applied for visualization purposes.}
  \label{fig:semi}
\end{figure}

Fig. \ref{fig:semi} compares SSTML with reference methods for real and semi-synthetic data streams.  As in the case of synthetic data, SSTML again achieves a BAC comparable to or exceeding reference methods, but some differences are also apparent. 

Lowered BAC value no longer characterizes SSTML at the beginning of the data stream,  which may result from the more heterogeneous data nature, which -- combined with STML encoding -- translates into a leveling of the playing field for SSTML and reference methods. In the case of synthetic data, which contained only informative features, its more homogeneous nature may have favored solutions based on classic Hoeffding trees. We also see that ROSE and KUE achieve better classification quality than in the case of synthetic streams, while NIE loses its properties. This is most likely due to the fact that as the heterogeneity of the tabular data increases, the mechanisms for dynamic weighting and selection of classifiers in KUE, as well as the techniques for ensuring ensemble diversity and drift detection in ROSE, gain utility. An interesting case is the poker stream, in which SSTML achieves a BAC value far superior to reference methods. This may be due to the relatively low number of features and their categorical nature.

\subsection{Experiment 3 --  Processing time}
The third experiment was conducted to investigate the processing time of SSTML compared to reference methods and to answer the third research question. For this purpose, synthetic data streams were generated with 110 data chunks, 250 samples each, and 8, 16, 32, and 64 features, respectively. The number of chunks was chosen so that it was possible to average the processing time from 100 batches, considering only those in which ensemble methods had already reached the maximum classifier pool size. Since the number of features of the problem is inherent to the size of the image in the STML representations (in order to ensure visibility of all features), the following quadrant image sizes were selected for the STML:8 features: 50 px, 16 features: 80 px, 32: 110 px, 64: 150 px.

\begin{figure}[!htb]
  \begin{center}
    \includegraphics[width=0.98\textwidth]{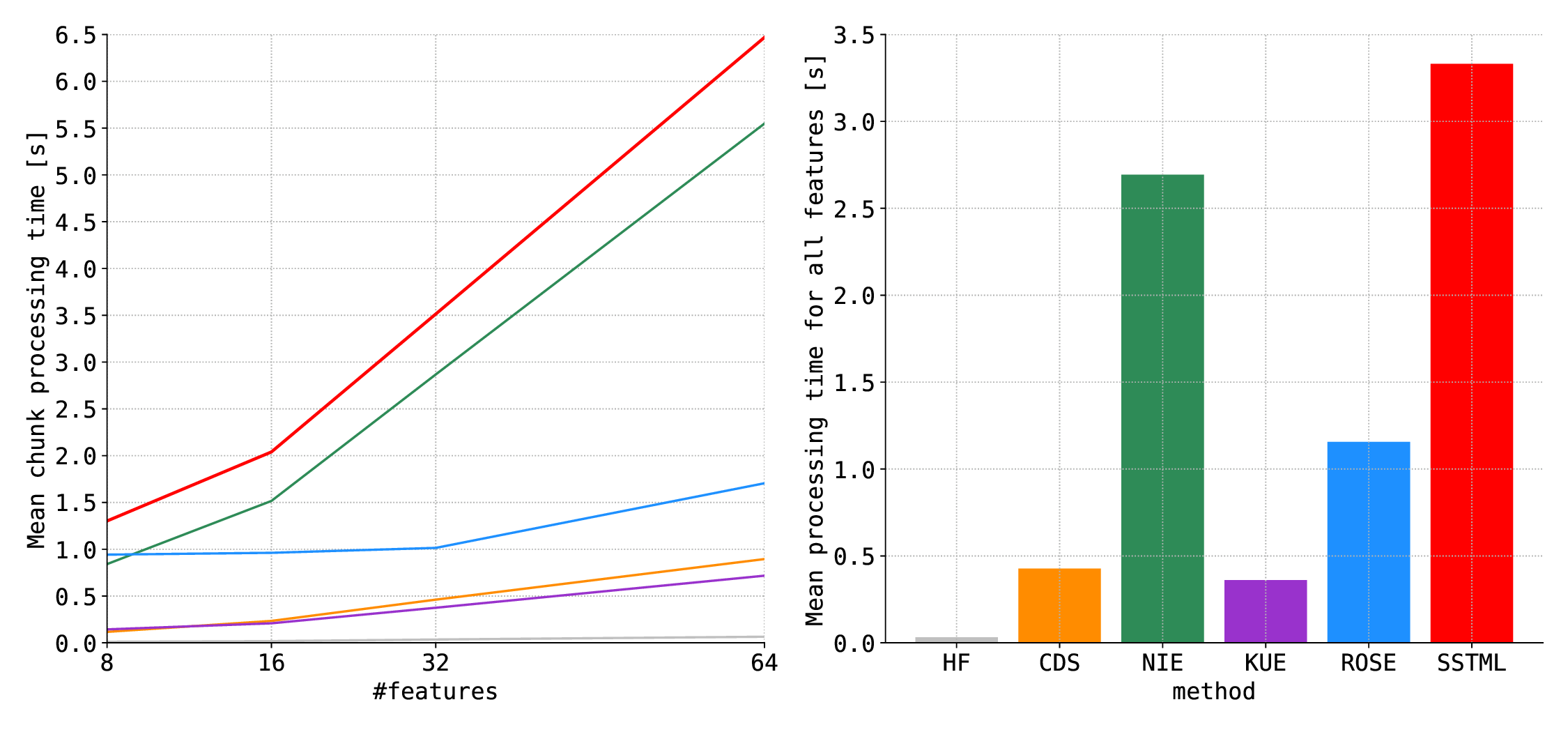}
  \end{center}
  \caption{Comparison of SSTML with reference methods in terms of processing time.}
  \label{fig:time}
\end{figure}

As we can see in Fig. \ref{fig:time}, SSTML, despite the need for a procedure to encode tabular data into a discrete digital signal and the use of the ResNet-18 architecture, does not have a processing time unprecedented among established ensemble algorithms, which are generally considered less complex. SSTML's average processing time is only about 0.5 seconds longer than NIE, which performs worse with synthetic and real streams. The popular ROSE algorithm offers processing times about 2.5 seconds faster than SSTML, but it is less universally applicable. In summary, SSTML, although it might seem otherwise based on the method description, does not offer processing times unheard of in currently used data stream classification algorithms and can be successfully used for low-velocity streams.

\section{Conclusion}
In an effort to fill a gap in the literature relating to the use of multi-dimensional encoding to classify difficult data streams, this paper proposes an SSTML approach. This is the first attempt known to the paper's authors to study the behavior of such methods when concept drift occurs. The results proved that SSTML can achieve statistically significantly better classification quality than state-of-the-art ensemble classification algorithms. Particularly evident is the higher generalization potential of SSTML, which is useful even when the learning phase is limited to one epoch. At the same time, the processing time of SSTML is not significantly different from state-of-the-art ensemble algorithms and can be used for batch-based processing of low-velocity data streams. All this shows that deep neural networks can be successfully used in stream data analysis instead of more classical approaches.

The results encourage further research into applying multi-dimensional encoding for difficult tabular data stream classification. Among the potential directions for development are an expanded analysis of available encoding methods and the use of different neural network architectures and loss functions to reduce processing time and potentially improve results.

\begin{credits}
\end{credits}

\bibliographystyle{splncs04}
\bibliography{bibliography}

\begin{thebibliography}{10}
\providecommand{\url}[1]{\texttt{#1}}
\providecommand{\urlprefix}{URL }
\providecommand{\doi}[1]{https://doi.org/#1}

\bibitem{aguiar2023survey}
Aguiar, G., Krawczyk, B., Cano, A.: A survey on learning from imbalanced data
  streams: taxonomy, challenges, empirical study, and reproducible experimental
  framework. Machine learning pp. 1--79 (2023)

\bibitem{aminian2019study}
Aminian, E., Ribeiro, R.P., Gama, J.: A study on imbalanced data streams. In:
  Joint European Conference on Machine Learning and Knowledge Discovery in
  Databases. pp. 380--389. Springer (2019)

\bibitem{bahri2021data}
Bahri, M., Bifet, A., Gama, J., Gomes, H.M., Maniu, S.: Data stream analysis:
  Foundations, major tasks and tools. Wiley Interdisciplinary Reviews: Data
  Mining and Knowledge Discovery  \textbf{11}(3),  e1405 (2021)

\bibitem{basu2022novel}
Basu, T., Menzer, O., Ward, J., SenGupta, I.: A novel implementation of siamese
  type neural networks in predicting rare fluctuations in financial time
  series. Risks  \textbf{10}(2), ~39 (2022)

\bibitem{batko2022use}
Batko, K., {\'S}l{\k{e}}zak, A.: The use of big data analytics in healthcare.
  Journal of big Data  \textbf{9}(1), ~3 (2022)

\bibitem{MOA}
Bifet, A., Holmes, G., Kirkby, R., Pfahringer, B.: {MOA:} massive online
  analysis. Journal of Machine Learning Research  \textbf{11},  1601--1604
  (2010)

\bibitem{borisov2022deep}
Borisov, V., Leemann, T., Se{\ss}ler, K., Haug, J., Pawelczyk, M., Kasneci, G.:
  Deep neural networks and tabular data: A survey. IEEE Transactions on Neural
  Networks and Learning Systems  (2022)

\bibitem{brzezinski2018ensemble}
Brzezinski, D., Stefanowski, J.: Ensemble classifiers for imbalanced and
  evolving data streams. In: Data mining in time series and streaming
  databases, pp. 44--68. World Scientific (2018)

\bibitem{cano2020kappa}
Cano, A., Krawczyk, B.: Kappa updated ensemble for drifting data stream mining.
  Machine Learning  \textbf{109}(1),  175--218 (2020)

\bibitem{cano2022rose}
Cano, A., Krawczyk, B.: Rose: robust online self-adjusting ensemble for
  continual learning on imbalanced drifting data streams. Machine Learning
  \textbf{111}(7),  2561--2599 (2022)

\bibitem{damri2023towards}
Damri, A., Last, M., Cohen, N.: Towards efficient image-based representation of
  tabular data. Neural Computing and Applications pp. 1--21 (2023)

\bibitem{Ditzler:2013}
Ditzler, G., Polikar, R.: Incremental learning of concept drift from streaming
  imbalanced data. IEEE Transactions on Knowledge and Data Engineering
  \textbf{25}(10),  2283--2301 (Oct 2013)

\bibitem{duda2020training}
Duda, P., Jaworski, M., Cader, A., Wang, L.: On training deep neural networks
  using a streaming approach. Journal of Artificial Intelligence and Soft
  Computing Research  \textbf{10}(1),  15--26 (2020)

\bibitem{gama2013evaluating}
Gama, J., Sebastiao, R., Rodrigues, P.P.: On evaluating stream learning
  algorithms. Machine learning  \textbf{90},  317--346 (2013)

\bibitem{guzy2020employing}
Guzy, F., Wo{\'z}niak, M.: Employing dropout regularization to classify
  recurring drifted data streams. In: 2020 International joint conference on
  neural networks (IJCNN). pp.~1--7. IEEE (2020)

\bibitem{haug2021learning}
Haug, J., Kasneci, G.: Learning parameter distributions to detect concept drift
  in data streams. In: 2020 25th international conference on pattern
  recognition (ICPR). pp. 9452--9459. IEEE (2021)

\bibitem{haug2020leveraging}
Haug, J., Pawelczyk, M., Broelemann, K., Kasneci, G.: Leveraging model inherent
  variable importance for stable online feature selection. In: Proceedings of
  the 26th ACM SIGKDD International Conference on Knowledge Discovery \& Data
  Mining. pp. 1478--1502 (2020)

\bibitem{he2023image}
He, Z., Sayadi, H.: Image-based zero-day malware detection in iomt devices: A
  hybrid ai-enabled method. In: 2023 24th International Symposium on Quality
  Electronic Design (ISQED). pp.~1--8. IEEE (2023)

\bibitem{kadra2021well}
Kadra, A., Lindauer, M., Hutter, F., Grabocka, J.: Well-tuned simple nets excel
  on tabular datasets. Advances in neural information processing systems
  \textbf{34},  23928--23941 (2021)

\bibitem{klikowski2020employing}
Klikowski, J., Wo{\'z}niak, M.: Employing one-class svm classifier ensemble for
  imbalanced data stream classification. In: Computational Science--ICCS 2020:
  20th International Conference, Amsterdam, The Netherlands, June 3--5, 2020,
  Proceedings, Part IV 20. pp. 117--127. Springer (2020)

\bibitem{krawczyk2017ensemble}
Krawczyk, B., Minku, L.L., Gama, J., Stefanowski, J., Wo{\'z}niak, M.: Ensemble
  learning for data stream analysis: A survey. Information Fusion  \textbf{37},
   132--156 (2017)

\bibitem{ksieniewicz2022stream}
Ksieniewicz, P., Zyblewski, P.: Stream-learn—open-source python library for
  difficult data stream batch analysis. Neurocomputing  \textbf{478},  11--21
  (2022)

\bibitem{leon2021dengue}
Leon, M.I., Iqbal, M.I., Meem, S., Alahi, F., Ahmed, M., Shatabda, S., Mukta,
  M.S.H.: Dengue outbreak prediction from weather aware data. In: International
  Conference on Bangabandhu and Digital Bangladesh. pp. 1--11. Springer (2021)

\bibitem{manapragada2018extremely}
Manapragada, C., Webb, G.I., Salehi, M.: Extremely fast decision tree. In:
  Proceedings of the 24th ACM SIGKDD International Conference on Knowledge
  Discovery \& Data Mining. pp. 1953--1962 (2018)

\bibitem{montiel2018scikit}
Montiel, J., Read, J., Bifet, A., Abdessalem, T.: Scikit-multiflow: A
  multi-output streaming framework. Journal of Machine Learning Research
  \textbf{19}(72), ~1--5 (2018)

\bibitem{sahoo2017online}
Sahoo, D., Pham, Q., Lu, J., Hoi, S.C.: Online deep learning: Learning deep
  neural networks on the fly. arXiv preprint arXiv:1711.03705  (2017)

\bibitem{satt2017efficient}
Satt, A., Rozenberg, S., Hoory, R., et~al.: Efficient emotion recognition from
  speech using deep learning on spectrograms. In: Interspeech. pp. 1089--1093
  (2017)

\bibitem{shwartz2022tabular}
Shwartz-Ziv, R., Armon, A.: Tabular data: Deep learning is not all you need.
  Information Fusion  \textbf{81},  84--90 (2022)

\bibitem{sun2019supertml}
Sun, B., Yang, L., Zhang, W., Lin, M., Dong, P., Young, C., Dong, J.: Supertml:
  Two-dimensional word embedding for the precognition on structured tabular
  data. In: Proceedings of the IEEE/CVF conference on computer vision and
  pattern recognition workshops. pp.~0--0 (2019)

\bibitem{Wang2015}
{Wang}, S., {Minku}, L.L., {Yao}, X.: Resampling-based ensemble methods for
  online class imbalance learning. IEEE Transactions on Knowledge and Data
  Engineering  \textbf{27}(5),  1356--1368 (May 2015)

\bibitem{wang2018systematic}
Wang, S., Minku, L.L., Yao, X.: A systematic study of online class imbalance
  learning with concept drift. IEEE transactions on neural networks and
  learning systems  \textbf{29}(10),  4802--4821 (2018)

\bibitem{wozniak2023active}
Wo{\'z}niak, M., Zyblewski, P., Ksieniewicz, P.: Active weighted aging ensemble
  for drifted data stream classification. Information Sciences  \textbf{630},
  286--304 (2023)

\bibitem{zhang2021neural}
Zhang, Q., Cao, L., Shi, C., Niu, Z.: Neural time-aware sequential
  recommendation by jointly modeling preference dynamics and explicit feature
  couplings. IEEE Transactions on Neural Networks and Learning Systems
  \textbf{33}(10),  5125--5137 (2021)

\bibitem{zhu2021converting}
Zhu, Y., Brettin, T., Xia, F., Partin, A., Shukla, M., Yoo, H., Evrard, Y.A.,
  Doroshow, J.H., Stevens, R.L.: Converting tabular data into images for deep
  learning with convolutional neural networks. Scientific reports
  \textbf{11}(1),  11325 (2021)

\bibitem{zhuang2020comprehensive}
Zhuang, F., Qi, Z., Duan, K., Xi, D., Zhu, Y., Zhu, H., Xiong, H., He, Q.: A
  comprehensive survey on transfer learning. Proceedings of the IEEE
  \textbf{109}(1),  43--76 (2020)

\bibitem{zyblewski2021preprocessed}
Zyblewski, P., Sabourin, R., Wo{\'z}niak, M.: Preprocessed dynamic classifier
  ensemble selection for highly imbalanced drifted data streams. Information
  Fusion  \textbf{66},  138--154 (2021)

\end{thebibliography}
\end{document}